\pdfoutput=1

\documentclass[11pt]{article}

\usepackage{emnlp2021}

\usepackage{times}
\usepackage{latexsym}

\usepackage[T1]{fontenc}

\usepackage[utf8]{inputenc}

\usepackage{microtype}

\usepackage{multirow}
\usepackage{soul}
\usepackage{graphicx}
\usepackage{microtype}
\usepackage{graphicx}
\usepackage{amsmath}
\usepackage{amsfonts}
\usepackage{amssymb}
\usepackage{textcomp}
\usepackage{xcolor}
\usepackage{cleveref}
\usepackage{mathrsfs}
\usepackage{amsthm}
\usepackage{bm}
\usepackage{booktabs}
\usepackage{enumitem}
\usepackage{times}
\usepackage{xspace}

\crefformat{section}{\S#2#1#3} 
\crefformat{subsection}{\S#2#1#3}
\crefformat{subsubsection}{\S#2#1#3}

\usepackage{CJKutf8}

\title{Multilingual AMR Parsing with Noisy Knowledge Distillation\thanks{~~This work was supported by Alibaba Group through the Alibaba Innovative Research (AIR) Program.}}
\author{Deng Cai$^\heartsuit$\quad Xin Li$^\spadesuit$\quad Jackie Chun-Sing Ho$^\heartsuit$\quad Lidong Bing$^\spadesuit$\quad Wai Lam$^\heartsuit$\\
	$^\heartsuit$The Chinese University of Hong Kong \\
	$^\spadesuit$DAMO Academy, Alibaba Group \\
	\texttt{thisisjcycd@gmail.com} \\
	\texttt{\{xinting.lx,l.bing\}@alibaba-inc.com} \\
	\texttt{jackieho@link.cuhk.edu.hk,wlam@se.cuhk.edu.hk}
	}
\begin{document}
	\maketitle
	\begin{abstract}
		We study multilingual AMR parsing from the perspective of knowledge distillation, where the aim is to learn and improve a multilingual AMR parser by using an existing English parser as its teacher. We constrain our exploration in a strict multilingual setting: there is but one model to parse all different languages including English. We identify that noisy input and precise output are the key to successful distillation. Together with extensive pre-training, we obtain an AMR parser whose performances surpass all previously published results on four different foreign languages, including German, Spanish, Italian, and Chinese, by large margins (up to 18.8 \textsc{Smatch} points on Chinese and on average 11.3 \textsc{Smatch} points). Our parser also achieves comparable performance on English to the latest state-of-the-art English-only parser.
	\end{abstract}
	\section{Introduction}
	Abstract Meaning Representation (AMR) \cite{banarescu-etal-2013-abstract} is a broad-coverage semantic formalism that encodes the meaning of a sentence as a rooted, directed, and labeled graph, where nodes represent concepts and edges represent relations among concepts. AMR parsing is the task of translating natural language sentences into their corresponding AMR graphs, which encompasses a set of natural language understanding tasks, such as named entity recognition, semantic role labeling, and coreference resolution. AMR has proved to be beneficial to a wide range of applications such as text summarization \cite{liao-etal-2018-abstract}, machine translation \cite{song-etal-2019-semantic}, and question answering \cite{kapanipathi2020question,xu-etal-2021-dynamic}.
	
\begin{figure}[t]
		\centering
		\includegraphics[scale=0.45]{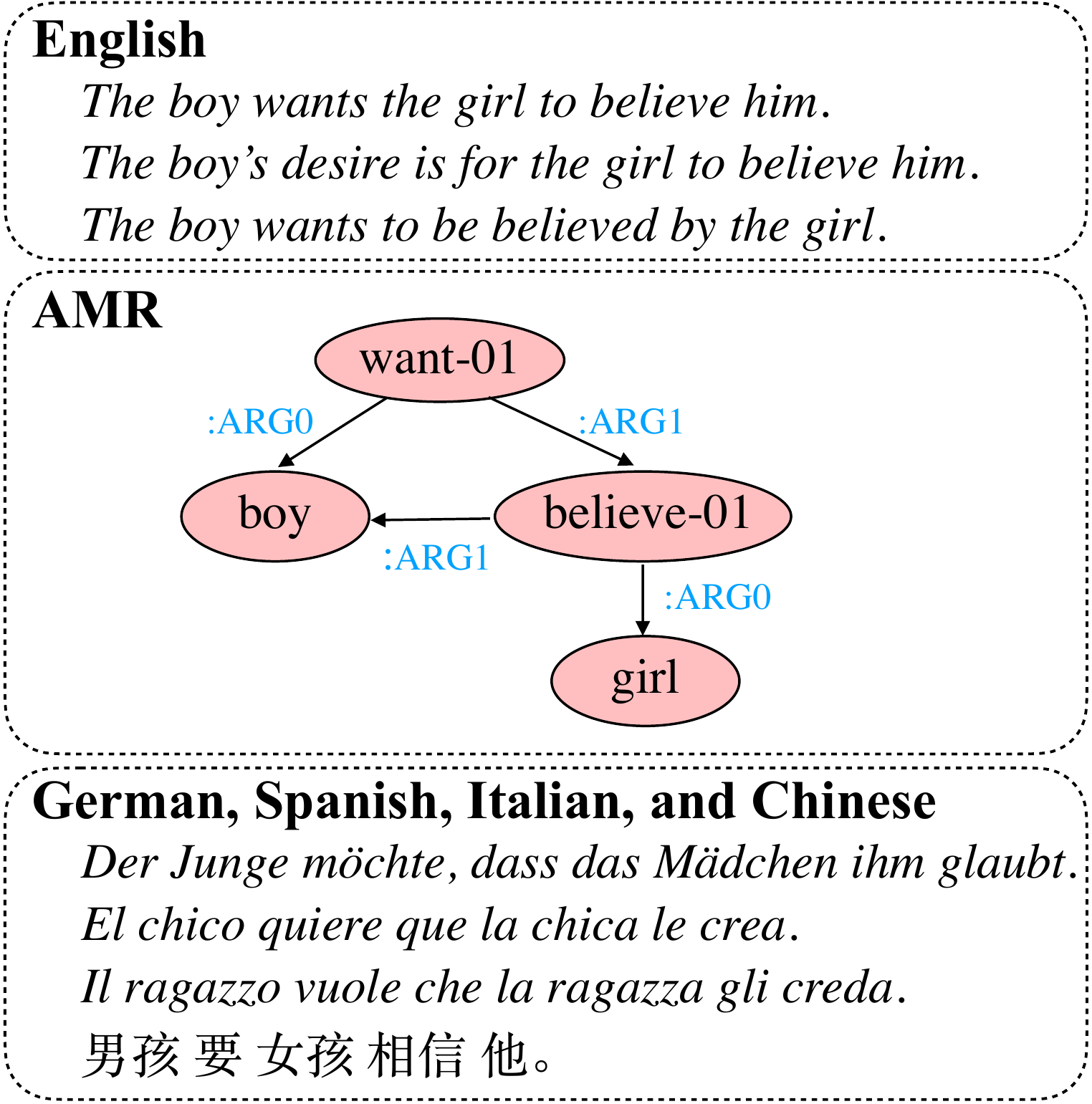}
		\caption{An example of AMR. Sentences written in English and other languages share the same meaning and therefore correspond to the same AMR graph.}
		\label{example-amr}
	\end{figure}

   One most critical feature of the AMR formalism is that it abstracts away from syntactic realization and surface forms. As shown in Figure \ref{example-amr}, different English sentences with the same meaning correspond to the same AMR graph. Furthermore, there are no explicit alignments between elements (nodes or edges) in the graph and words in the text. While this property leads to a distinct difficulty in AMR parsing, it also suggests the potential of AMR to work as an interlingua \cite{xue-etal-2014-interlingua,hajic-etal-2014-comparing,damonte-cohen-2018-cross}, which could be useful to multilingual applications of natural language understanding \cite{liang2020xglue,hu2020xtreme}. An example is given in Figure \ref{example-amr}, we represent the semantics of semantically-equivalent sentences in other languages using the same AMR graph. This defines the multilingual AMR parsing problem we seek to address in this paper. 

   Multilingual AMR parsing is an extremely challenging task due to several reasons. First, AMR was initially designed for and heavily biased towards English, thus the parsing has to overcome some structural linguistic divergences among languages \cite{damonte-cohen-2018-cross,zhu-etal-2019-towards}. Second, the human-annotated resources for training are only available in English and none is present in other languages. Moreover, since the AMR graph involves rich semantic labels, the AMR annotation for other languages can be labor-intensive and unaffordable. Third, current modeling techniques focus mostly on English. For example, existing AMR aligners \cite{flanigan-etal-2014-discriminative,pourdamghani-etal-2014-aligning,liu-etal-2018-amr} and widely-used pointer-generator mechanisms \cite{zhang-etal-2019-broad,cai-lam-2019-core,cai-lam-2020-amr} rely on the textual overlap between English words and AMR node values (i.e., concepts).
   
   Some initial attempts \cite{damonte-cohen-2018-cross,blloshmi-etal-2020-xl,sheth2021bootstrapping} towards multilingual AMR parsing mainly investigated the construction of pseudo parallel data via annotation projection. In this paper, we study multilingual AMR parsing from the perspective of knowledge distillation \cite{bucilua2006model,ba2014deep,hinton2015distilling,kim-rush-2016-sequence}, where our primary goal is to improve a multilingual AMR parser by using an existing English parser as its teacher. We focus on a strict multilingual setting for developing one AMR parser that can parse all different languages. In contrast to the language-specific (one parser one language) setting, our setting is more challenging yet more appealing in practice. Intuitively, knowledge distillation is effective because the teacher’s output provides a rich training signal for the student parser. We develop both the teacher parser and the student parser with language-agnostic \textit{seq2seq} design and expect the student parser to imitate the behaviors of the teacher parser (i.e., English parser) when processing semantically-equivalent input in other languages. We first show that multilingual \textit{seq2seq} pre-training, including language model and machine translation pre-training, provides an excellent starting point for model generalization across languages. We further capitalize on the idea that the student should be robust to noisy input and introduce noise by machine translation for improving student performance. To mitigate the risk that the student learns the mistakes made by the teacher, the student is then fine-tuned with gold AMR graphs. 
	
   We present experiments on the benchmark dataset created by \newcite{damonte-cohen-2018-cross}, covering four different languages with no training data, including German, Spanish, Italian, and Chinese. To cover as many languages as possible, we also include the original English test set in our evaluation. On four zero-resource languages, our single universal parser consistently outperforms the previous best results by large margins (+11.3 \textsc{Smatch} points on average and up to +18.8 \textsc{Smatch} points). Meanwhile, our parser achieves competitive results on English even compared with the latest state-of-the-art English AMR parser in the literature.
	
	To sum up, our contributions are listed below:
	\begin{itemize}
	\item We study AMR parsing in a strict multilingual setting, there is but one parser for all different languages including English.
	\item We propose to train a multilingual AMR parser with multiple pre-training and fine-tuning stages including noisy knowledge distillation.
	\item We obtain a performant multilingual AMR parser, establishing new state-of-the-art results on multiple languages. We hope our parser can facilitate the multilingual applications of AMR. 
	\end{itemize}
	\section{Background}
	\subsection{Prior Work}
	Cross-lingual AMR parsing is the task of mapping a sentence in any language X to the AMR graph of its English translation. To date, there is no human-annotated X-AMR parallel dataset for training. Therefore, one straightforward solution is to translate the sentences from X into English then apply an English parser \cite{damonte-cohen-2018-cross,uhrig2021translate}. However, it is argued that the method is not informative in terms of the cross-lingual properties of AMR \cite{damonte-cohen-2018-cross,blloshmi-etal-2020-xl}. To tackle cross-lingual AMR parsing, most previous work relies on pre-trained multilingual language models and silver training data (i.e., pseudo parallel data).
	
	\paragraph{Pre-trained Multilingual Language Model}
	Previous work proves that language-independent features provided by pre-trained multilingual language models can boost cross-lingual parsing performance. For example, \newcite{blloshmi-etal-2020-xl} use mBERT \cite{devlin-etal-2019-bert} and \newcite{sheth2021bootstrapping} employ XLM-R \cite{conneau-etal-2020-unsupervised}.
	\paragraph{Silver Training Data}
	There are two typical methods for creating silver training examples: (\textbf{I}) \textit{Parsing English to AMR} \cite{damonte-cohen-2018-cross}. This approach creates silver training examples for the foreign language X through an external X-EN parallel corpus and an existing English AMR parser. The English sentences of the parallel corpus are parsed using the existing AMR parser. Then resultant AMR graphs are used as pseudo targets. Note that the target side of the constructed X-AMR training corpus is of silver quality. (\textbf{II}) \textit{Translating English to X} \cite{blloshmi-etal-2020-xl,sheth2021bootstrapping}. This approach does not exploit external X-EN parallel corpus but makes use of the existing EN-AMR parallel corpus. It uses off-the-shelf machine translation systems to translate the English side of the EN-AMR pairs into the foreign language X. It is worth noting that although the source side of the constructed training examples may contain noise introduced by automatic translation, the target side consists of gold AMR graphs. However, the number of training examples created by this approach is limited by the size of the original English dataset.
	\subsection{Our Task: Multilingual AMR Parsing}
	Here, we formally define the task of multilingual AMR parsing. As illustrated in Figure \ref{example-amr}, this task aims to predict the semantic graph given the input sentence in any language. Specifically, we consider five different languages: German (DE), Spanish (ES), Italian (IS), Chinese (ZH), and English (EN). The biggest challenge is due to the only access to a set of human-annotated English training examples. Formally, denote $\mathbb{Z}=\{\text{DE}, \text{ES}, \text{IT}, \text{ZH}\}$ as the set of \textit{foreign languages} other than EN. Our goal is to develop a multilingual parser for all languages in $\{\text{EN}\}\cup\mathbb{Z}$. However, there only exists a set of gold EN-AMR training pairs $(x, y)$ where $x$ and $y$ are the English sentence and AMR graph, respectively. For any language X$\in\mathbb{Z}$, there is \textit{no} gold training example.

	Following the recent state-of-the-art practice for English AMR parsing \cite{bevilacqua-etal-2021-one}, we formulate the problem as a \textit{seq2seq} task. The input sentence serves as the source sequence, while the linearization of the AMR graph is treated as the target sequence.
	\section{Our Parser}
	\subsection{Overview}
	We choose vanilla \textit{seq2seq} architecture \cite{vaswani2017attention,bevilacqua-etal-2021-one} for our multilingual AMR parser to dispose of the need of explicit word-to-node alignments. Unlike \newcite{damonte-cohen-2018-cross,sheth2021bootstrapping}, the advantage of alignment-free parsers is that the training is prevented from depending on noisy alignments derived from automatic cross-lingual aligners.
	
	The training of our parser consists of multiple pre-training and fine-tuning stages. First, we initialize both the encoder and decoder of our parser using parameters pre-trained for multilingual denoising autoencoding and multilingual machine translation. We argue that both pre-training stages boost model generalization across languages and the latter is especially beneficial to AMR parsing because translating to a meaning representation resembles machine translation. Then, we fine-tune our parser in two stages. In the first stage, we aim to transfer the knowledge of a high-performing English AMR parser to our multilingual parser via knowledge distillation. Finally, we fine-tune our parser with gold AMR graphs to alleviate the drawback of over-fitting to teacher's mistakes. Each training stage is detailed in \cref{training-stages} and its individual effect is empirically revealed in \cref{main-results}.
	\begin{figure*}[t]
		\centering
		\includegraphics[scale=0.44]{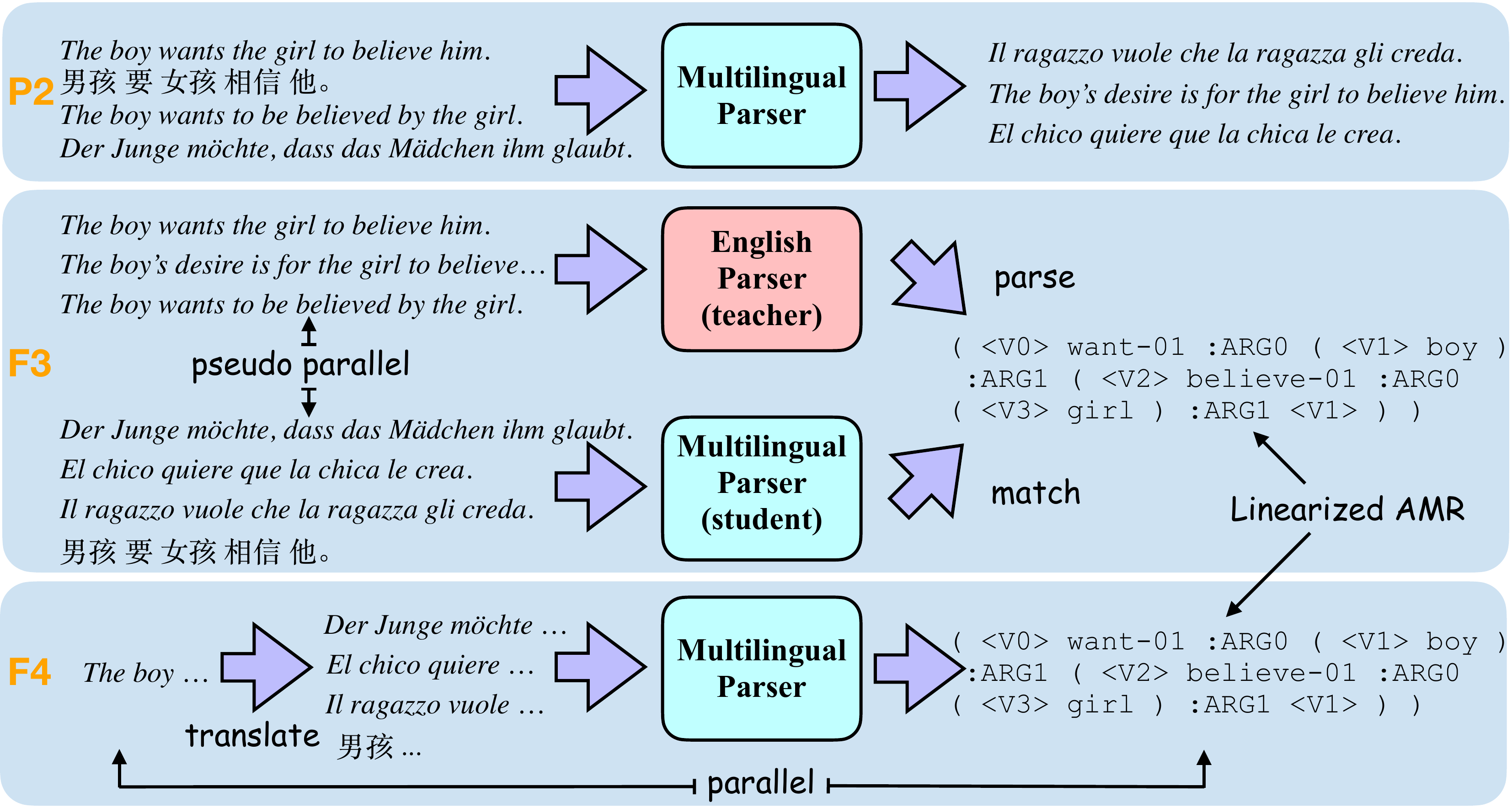}
		\caption{Illustration of different training stages. Stage P1 is omitted for space limit.}
		\label{fig-training-stages}
	\end{figure*}
	\subsection{Base Model}
	\label{base-model}
	\paragraph{Model Architecture} We consider the standard \textit{Transformer} \cite{vaswani2017attention} for \textit{seq2seq} modeling. The encoder in the Transformer consists of a stack of multiple identical layers, each of which has two sub-layers: one implements the multi-head self-attention mechanism and the other is a position-wise fully connected feed-forward network. The decoder is also composed of a stack of multiple identical layers. Each layer in the decoder consists of the same sub-layers as in the encoder layers plus an additional sub-layer that performs multi-head attention to the output of the encoder stack. See \newcite{vaswani2017attention} for more details.
	
	\paragraph{Linearization \& Post-processing}
	To formulate AMR parsing as a \textit{seq2seq} problem, one needs to first obtain the linearized sequence representation of AMR graphs. To this end, we adopt the fully graph-isomorphic linearization techniques as in \newcite{bevilacqua-etal-2021-one}. That is, the graph is recoverable from the linearized sequence without losing adjacency information. We use special tokens $\texttt{<V0>, <V1>, \ldots, <Vn>}$ to represent variables in the linearized graph and to handle co-referring nodes. We make a clear distinction between constants and variables, as variable names do not carry any semantic information. The graph is linearized through a depth-first traversal starting from the root. For edge ordering, we use the default order in the release files of AMR datasets as suggested by \newcite{konstas-etal-2017-neural}. The bottom right of Figure \ref{fig-training-stages} illustrates the linearization result of the AMR graph in Figure \ref{example-amr}.
	
	The output sequence of our \textit{seq2seq} model may produce an invalid graph. For example, the parenthesis parity may be broken, resulting in an incomplete graph. To ensure the validity of the graph produced in parsing, post-processing steps such as parenthesis parity restoration and invalid segment removal are introduced. We use the pre- and post-processing scripts provided by \newcite{bevilacqua-etal-2021-one}.\footnote{\url{https://github.com/SapienzaNLP/spring}}
	\subsection{Training Stages}
	\label{training-stages}
	We now clarify the four different training stages. The whole training process is referred to as P1$\rightarrow$P2$ \rightarrow$F3$\rightarrow$F4.
	\paragraph{P1: Multilingual Language Model Pre-training}
    Pre-trained multilingual language representations such as mBERT \cite{devlin-etal-2019-bert} have greatly improved performance across many cross-lingual language understanding tasks. For cross-lingual AMR parsing, in particular, \newcite{blloshmi-etal-2020-xl} used mBERT\footnote{\texttt{bert-base-multilingual-cased}} \cite{devlin-etal-2019-bert} while \newcite{sheth2021bootstrapping} employed XLM-R\footnote{\texttt{xlm-roberta-large}} \cite{conneau-etal-2020-unsupervised} to provide language-independent features. Unlike previous work, we argue that such encoder-only pre-trained models are not the most suitable choice for our \textit{seq2seq} parser. Instead, we adopt mBART, an encoder-decoder denoising language model pre-trained with monolingual corpora in many languages \cite{liu-etal-2020-multilingual}, to initialize both the encoder and decoder of our \textit{seq2seq} parser.
    \paragraph{P2: Multilingual Machine Translation Pre-training (MMT-PT)}
    The task of multilingual machine translation (MMT) is to learn one single model to translate between various language pairs. Essentially, natural languages can be considered as \textit{informal} meaning representations compared to \textit{formal} meaning representation such as AMR. On the other hand, AMR can be regarded as a special language. The above observations connect the dots between MMT and multilingual AMR parsing, both of which model the process of digesting the semantics in one form and and conveying the same semantics in another form. Therefore, we argue that pre-training our parser using the MMT task should be helpful. In fact, the usefulness of MT pre-training has also been validated in English AMR parsing \cite{xu-etal-2020-improving}. In practice, we directly use the mBARTmmt checkpoint \cite{tang2020multilingual}, an MMT model covering 50 languages that are trained from mBART.
	\paragraph{F3: Knowledge Distillation Fine-tuning (KD-FT)}
	Motivated by the fact that the parsing accuracy on English is significantly better than those on other languages, we propose to reduce the performance gap via knowledge distillation \cite{kim-rush-2016-sequence}. Specifically, we first pre-train a high-performance AMR parser for English and treat it as the teacher model. By considering our multilingual AMR parser as the student model, the goal is to transfer the knowledge of the teacher model to the student model. Intuitively, when feeding an English sentence to the teacher model, the student model, which receives its translation as input, should imitate the behaviors of the teacher model and make similar predictions. The strategies we adopt to achieve this goal are detailed in \cref{noisy-kd}.
	\paragraph{F4: Gold AMR Graph Fine-tuning (Gold-FT)}
	Note that in the knowledge distillation stage, the student parser is trained to match the predictions of the teacher model. A potential risk of such knowledge distillation is that the mistakes made by the teacher model may be propagated to the student model as well. Another fine-tuning stage, which we found useful for alleviating the risk, is to further fine-tune our parser with gold AMR graphs. Following \newcite{blloshmi-etal-2020-xl,sheth2021bootstrapping}, we transform the gold standard English datasets into other languages using MT models for fine-tuning our multilingual AMR parser.
	\subsection{Knowledge Distillation}
	\label{noisy-kd}
	Knowledge distillation (KD) refers to a class of techniques for training a \textit{student} model to imitate a \textit{teacher} model for close or even better performance. In contrast to most KD applications that focus on reducing the performance gap caused by architectural differences, our primary goal is to minimize the mismatch of model behaviors across languages. That is, we expect the student and teacher to behave similarly even with \textit{different} input languages.
	
	Recall that we formulate AMR parsing as a \textit{seq2seq} problem with standard maximum likelihood estimation training objective.
	\begin{equation}
	    L_{\text{MLE}} = \sum_{t=1}^{|y|} \log p(y_t|y_{:<t},x)
	    \label{mle}
	\end{equation}
	where $y_t$ denotes the $t$-th token in the linearized AMR sequence. One natural and common method for KD is to replace the discrete target with the soft token-level distributions provided by the teacher model $p_{T}(y_t|y_{:<t},x^*)$.
	\begin{equation}
	    L_{\text{token}} = \sum_{t=1}^{|y|} \text{KL}((p(y_t|y_{:<t},x), p_{T}(y_t|y_{:<t},x^*))
	    \nonumber
	\end{equation}
	where KL computes the Kullback–Leibler divergence between two distributions. We use $x^*$ and $x$ to highlight that the input sentences are in different languages. The above method is referred to as \textit{token-level} KD as it attempts to match the local token distributions of the teacher model. Opposed to token-level KD, \textit{sequence-level} KD \cite{kim-rush-2016-sequence} allows knowledge transfer at sequence-level $L_{\text{seq}} =  \text{KL}(p(y,x),p_{T}(y,x^*))$. Due to the intractability of sequence-level distribution computation, following \newcite{kim-rush-2016-sequence}, we replace the teacher's distribution with its mode. Specifically, we use beam search to approximate the teacher's most probable output, which is then used as the target to train the student model as in Eq. \ref{mle}.
	
	One appealing property of sequence-level KD is that it does not require gold AMR graphs. Therefore, it can be performed with an external X-EN parallel corpus at scale. However, the inherent noise in the teacher's output hampers training with the student often being prone to hallucination \cite{liu2020noisy}. To alleviate this problem, we propose to also inject noise to the input side of the student model. We find that automatic translation can serve as an effective noise generator for multilingual AMR parsing. That is, instead of using gold translations, we feed automatic machine translations to the student model. We find that the noise introduced by machine translation performs better than random noise likely due to that the translations preserve the most salient semantics.
	\section{Experimental Setup}
	\subsection{Datasets}
	\paragraph{Gold Data}
	Following conventions, we use the benchmark dataset created in \newcite{damonte-cohen-2018-cross} as our testbed. This dataset contains human translations of the test set of AMR2.0 dataset (LDC2017T10) in German (DE), Spanish (ES), Italian (IT), and Chinese (ZH). For a more complete multilingual setup, we also include the original English (EN) test set for evaluation. The gold training corpus in our experiments is the training set of AMR2.0, which contains 36, 521 EN-AMR pairs.
	\paragraph{Silver Data}
	For other foreign languages (DE, ES, IT, and ZH), we construct silver training data following \newcite{blloshmi-etal-2020-xl}. Specifically, we use OPUS-MT \cite{tiedemann-thottingal-2020-opus}\footnote{\url{https://huggingface.co/transformers/model_doc/marian.html}}, an off-the-shelf translation tool, to translate English sentences in AMR2.0 to other foreign languages. To ensure the quality of silver data, we filter out data with less accurate translations via back-translation consistency check. That is, the translation quality is measured by the cosine similarity between the original English sentence and its back-translated counterpart using LASER \cite{artetxe-schwenk-2019-massively}. We refer readers to \newcite{blloshmi-etal-2020-xl} for an exhaustive description of the data filtering process. Detailed statistics of our training, dev, and test sets are shown in Table \ref{tb-data}.
	\paragraph{Knowledge Distillation Data}
	For the knowledge distillation stage, we use 320K English sentences in the Europarl corpus \cite{koehn2005europarl}, which contains parallel sentence pairs of En$\Leftrightarrow$DE, En$\Leftrightarrow$ES, and En$\Leftrightarrow$IT. Unless otherwise specified, we use sequence-level KD with noisy input from OPUS-MT. Note that essentially our noisy KD only requires monolingual English data. Nevertheless, we choose Europarl following \newcite{damonte-cohen-2018-cross,blloshmi-etal-2020-xl} and use the gold translations as noise-free input to demonstrate the impact of our noisy KD comparatively (\cref{discussions}).
	\subsection{Settings} We differentiate two settings for training and evaluating multilingual AMR parsing. 
	\begin{itemize}
		\item \textbf{Language-specific}. For each target language, a language-specific parser is trained.
		\item \textbf{Multilingual}. One single parser is trained to parse all target languages.
	\end{itemize}
	While this paper focuses on the multilingual setting, we also report the results of the language-specific parsers in previous work \cite{damonte-cohen-2018-cross,blloshmi-etal-2020-xl,sheth2021bootstrapping} for comparative reference.
	\subsection{Models}
	\paragraph{Model Variants}
	Our full training pipeline consists of multiple pre-training and fine-tuning stages. To study the effect of each training stage, we implement a series of model variants:
   \begin{itemize}
       \item \textbf{w/o MMT-PT}. To measure the help from MMT-PT, we remove the second pre-training stage (P2). The training process becomes P1$\rightarrow$F3$\rightarrow$F4. 
       \item \textbf{w/o KD-FT}. To show the benefits from KD, we conduct an ablation experiment where the KD-FT stage (F3) is skipped. The training process becomes P1$\rightarrow$P2$\rightarrow$F4.
       \item \textbf{w/o Gold-FT}. To validate the necessity of the fine-tuning with gold AMR graph, we also report the model results without the final Gold-FT (F4) stage. The training process is then P1$\rightarrow$P2$\rightarrow$F3.
       \item \textbf{w/o MMT-PT \& KD-FT}. We exclude both the MMT-PT (P2) stage and the KD-FT (F3) stage. This variant (P1$\rightarrow$F4) is reminiscent of the best-performing model of \newcite{blloshmi-etal-2020-xl} that fine-tunes multilingual language model with silver training data.
       \item \textbf{w/o MMT-PT \& Gold-FT}. We also report the model performance without MMT-PT and Gold-FT for reference (P1$\rightarrow$F3).
   \end{itemize}
   		    \begin{table}
    	\small
		\centering
		\begin{tabular}{llll}
			\hline
			
			\hline
		      Language & Train & Dev & Test \\
		    \hline
		     English(EN)& 36,521$^*$ & 1,368$^*$ & 1,371$^*$ \\
		     German(DE) & 34,415 & 1,319 & 1,371$^*$ \\
		     Spanish(ES) & 34,552 & 1,325 & 1,371$^*$ \\
		     Italian(IT) & 34,521 & 1,322 & 1,371$^*$ \\
		     Chinese(ZH) & 33,221 & 1,311 & 1,371$^*$ \\
			\hline
			
			\hline
		\end{tabular}
		\caption{The number of instances per language and for each data split. $^*$ marks gold quality and otherwise silver quality.}
		\label{tb-data}
	\end{table}
   					\begin{table*}
		\centering
		\small
		\begin{tabular}{lrccccccc}
			\hline
			
			\hline
			\multirow{2}{*}{Model}& & \multicolumn{7}{c}{\textsc{Smatch}} \\
			& & DE & ES & IT & ZH & EN & AVG$_{\text{X}}$ & AVG \\
			\hline
		
		\multicolumn{9}{l}{\textbf{Language-Specific}}\\
			~\newcite{damonte-cohen-2018-cross} & &39.0&42.0&43.0&35.0& - &39.8&- \\
			~\newcite{blloshmi-etal-2020-xl} && 53.0 & 58.0 & 58.1 & 43.1 &-&53.1&-\\
			~\newcite{sheth2021bootstrapping}$\dagger$ && 62.7 & 67.9 & 67.4 &-&-&-&- \\
			\hline
			\multicolumn{9}{l}{\textbf{Multilingual}} \\
			~\newcite{blloshmi-etal-2020-xl}$\dagger$ && 52.1&56.2&56.7 & - &-&-&-\\
			~\newcite{blloshmi-etal-2020-xl} && 49.9&53.2&53.5 & 41.0 & -& 49.4&-\\
			~Ours&\small{(P1$\rightarrow$P2$
			\rightarrow$F3$\rightarrow$F4)} &\textbf{73.1} &	\textbf{75.9}&	\textbf{75.4}&	\textbf{61.9}&\textbf{83.9}&	\textbf{71.6}&	\textbf{74.0}\\
			~~w/o MMT-PT &\small{(P1$\rightarrow$F3$\rightarrow$F4)} &72.4&	75.6&	\textbf{75.4}&	60.6&	83.3&	71.0&	73.5\\
			~~w/o KD-FT & \small{(P1$\rightarrow$P2$\rightarrow$F4)}  &71.8&	74.5&	73.8&	61.0&	82.6&	70.3&	72.7\\
			~~w/o Gold-FT & \small{(P1$\rightarrow$P2$\rightarrow$F3)}  &70.9&	74.0&	73.1&	59.5&	82.4&	69.4&	72.0\\
			~~w/o MMT-PT \& KD-FT & \small{(P1$\rightarrow$F4)}  &70.8&	73.8&	73.2&	59.9&	81.8&	69.4&	71.9\\
			~~w/o MMT-PT \& Gold-FT & \small{(P1$\rightarrow$F3)} &70.0&	73.3&	72.7&	58.4&	81.4&	68.6&	71.2\\
			\hline
			\multicolumn{9}{l}{\textbf{State-of-the-art English-only Parser}}\\
			\newcite{bevilacqua-etal-2021-one}$\ddagger$ &&-&-&-&-&84.3&-&-\\
			\hline
			
			\hline
		\end{tabular}
		\caption{\textsc{Smatch} scores on test sets. AVG$_{\text{X}}$ and AVG denote the averages over zero-resource languages (DE, ES IT, and ZH) and all languages respectively. $\dagger$ indicates that the results do not include ZH. $\ddagger$ marks that we report the best score without graph re-categorization considering our models do not use graph re-categorization either.\footnotemark[7]}
		\label{results}
	\end{table*}
	\paragraph{Implementation Details}
	Following \newcite{bevilacqua-etal-2021-one}, we make slight modifications to the vocabulary of mBART for better suiting linearized AMRs. Specifically, we augment the original vocabulary of mBART with the names of AMR relations and frames occurring at least 5 times in the gold training corpus. The augmented vocabulary allows more compact target sequence after tokenization. As introduced in \cref{training-stages}, the first two pre-training stages are out of scope for this paper and we directly load pre-trained model checkpoints, mBART\footnote{\texttt{{facebook/mbart-large-50}}} \cite{liu-etal-2020-multilingual} and mBARTmmt\footnote{\texttt{facebook/mbart-large-50-{\scriptsize many-to-many}-mmt}} \cite{tang2020multilingual}, from Huggingface’s transformers library \cite{wolf-etal-2020-transformers}. At each fine-tuning stage, models are trained for up to 30,000 steps with a batch size of 5,000 graph linearization tokens, with RAdam \cite{liu2019variance} optimizer and a learning rate of 1e-5. Dropout is set to 0.25. We do model selection according to the performance on dev sets. At prediction time, we set beam size to 5. The teacher model is separately trained and obtains 84.2 \textsc{Smatch} score on the English test set, which is close to the recent state-of-the-art result \cite{bevilacqua-etal-2021-one}. We release our code, data, and models at \url{https://github.com/jcyk/XAMR}.
	\section{Experimental Results}
	\footnotetext[7]{Graph re-categorization is a popular technique for reducing the complexity of AMR graphs, which involves manual efforts for hand-crafting rules. Recent work \cite{bevilacqua-etal-2021-one} points out that graph re-categorization may harm the generalization ability to out-of-domain data.}
	The performance of AMR parsing is conventionally measured by \textsc{Smatch} score \cite{cai-knight-2013-smatch}, which quantifies the maximum overlap between two AMR graphs. The reported results are averaged over 3 runs with different random seeds.
	\subsection{Main Results}
	\label{main-results}
	In Table \ref{results}, we present the \textsc{Smatch} scores of our models and the best-performing models in the current literature. Our model with the full training pipeline achieves new state-of-the-art performances on all the four zero-resource languages, substantially outperforming all previous results. Concretely, the performance gains over the previous best results \cite{sheth2021bootstrapping} are 10.4, 8.0, 8.0, and 18.8 \textsc{Smatch} points on German, Spanish, Italian, and Chinese respectively. This is even more remarkable given that the previous best results are achieved via a set of language-specific parsers, while ours are obtained by one single multilingual parser. Notably, our multilingual parser also obtains close performance on English to that achieved by the state-of-the-art English-only parser. These results are encouraging for developing AMR parser in a strict multilingual setting (i.e., using one parser for all languages).
	
	The results of our ablated model variants further reveal the source of performance gains. As seen, each of MMT-PT, KD-FT, and Gold-FT make indispensable contributions to the superior performance. Skipping any of them leads to a considerable performance drop and removing two further degrades the model performance. Concretely, the averaged \textsc{Smatch} score across all languages (AVG) decreases by 0.5 points when removing MMT-PT, which confirms our hypothesis that MMT is a beneficial pre-training objective for multilingual AMR parsing. It is also observed that the AVG score drops down from 74.0 to 72.7 ($-$1.3 points) when skipping KD-FT. In other words, introducing KD-FT boosts the performance by 1.3 \textsc{Smatch} points on average. The improvement is striking since \textbf{Ours w/o KD-FT} is already a very strong baseline (AVG$=$72.7). Lastly, by comparing the results of \textbf{Ours w/o Gold FT} and \textbf{Ours}, we can see that appending Gold-FT to the preceding training stages yields a growth of 2.0 AVG points. This demonstrates that KD-FT alone is not sufficient and fine-tuning with gold AMR graphs has a complementary effect. Another interesting finding is that even our worst-performing variant surpasses previous best methods, which validates that pre-trained encoder-decoder architecture, mBART, is more effective for multilingual AMR parsing than encoder-only pre-trained models used in prior work.
	\subsection{Discussions}
	\label{discussions}
	Now we delve into more discussions on our key innovation, i.e., the knowledge distillation stage.
	\paragraph{Effect of Different Knowledge Distillation Methods}
	As introduced in \cref{noisy-kd}, there are two kinds of knowledge distillation (KD) methods for \textit{seq2seq} tasks: token-level KD (\textit{tok}) and sequence-level KD (\textit{seq}). In Table \ref{res-kd}, we compare \textit{tok}, \textit{seq}, and their combination (\textit{seq}+\textit{tok}). For \textit{seq}+\textit{tok}, we train the student on teacher-generated graphs but still use a token-level KL term between the teacher/student. Note that \textit{tok} can only utilize data with gold AMR graphs (i.e., the constructed silver training data), while \textit{seq} and \textit{seq}+\textit{tok} leverage additional English sentences. Therefore, we also report the result of \textit{seq} using the same English sentences as \textit{tok}, denoted as \textit{seq}$^*$. As seen, \textit{seq} performs much better than \textit{tok} and their combination does not bring further improvement. However, \textit{seq}$^*$ only gives similar result to \textit{tok}. These results show that training on more data is crucial and using \textit{seq} alone is sufficient for knowledge transfer.
	 
    \paragraph{Effect of Noise for Knowledge Distillation}
    Next, we study the effect of noise during knowledge distillation. Recall that we use automatic machine translation to generate noisy input for the student model. To show that noise is an important ingredient for superior performance, we also conduct experiments where the reference translations in Europarl are used as noise-free input to the student. Also, to show that the noise from MT is non-trivial, we further employ BART-style random noise \cite{lewis-etal-2020-bart} for comparison. BART-style noise masks text spans in the input and we tune the rate of word deletion. The results are presented in Table \ref{res-noise}. We show that MT noise is indeed helpful and its role cannot be replaced by simple random noise.
    	    \begin{table}
		\centering
		\small
		\begin{tabular}{lcccccc}
			\hline
			
			\hline
			Method & DE & ES & IT & ZH & EN & AVG \\
            \textit{tok}         &71.8	 &75.1	 &74.0	 &60.9	 &82.7	 &72.9 \\
            \textit{seq}        &73.1	 &75.9	 &75.4 &61.9	 &83.9	 &74.0 \\
            \textit{tok} + \textit{seq}  &73.1	 &75.8	 &75.3	 &61.6	 &83.9	 &73.9 \\
            \hline
            \textit{seq}$^*$           &71.9	 &75.0	 &74.1	 &61.2	 &82.9	 &73.0 \\
			\hline
			
			\hline
		\end{tabular}
		\caption{Comparison of different KD methods.}
		\label{res-kd}
	\end{table}
        \begin{table}
		\centering
	    \small
		\begin{tabular}{lcccccc}
			\hline
			
			\hline
		      Noise & DE & ES & IT & ZH & EN & AVG \\
		    \hline
		    None & 72.3	&75.3	&74.8	&61.3	&83.1	&73.4 \\
		    \hline
		     \multicolumn{7}{l}{Word deletion}\\
		     ~10\% & 72.4&	75.1&	74.6&	61.3&	83.5&	73.4\\
		     ~15\% & 72.4&	75.1&	74.7&	61.6&	83.3&	73.4 \\
		     ~20\% & 72.7&	75.6&	75.1&	61.3&	83.7&	73.7 \\
		     ~25\% & 72.5&	75.2&	74.3&	61.1&	83.3&	73.3 \\
		     ~30\% & 72.5&	75.3&	74.7&	61.3&	83.5&	73.4 \\
		   \hline
		     MT &\textbf{73.1} &	\textbf{75.9}&	\textbf{75.4}&	\textbf{61.9}&\textbf{83.9}&	\textbf{74.0}  \\
			\hline
			
			\hline
		\end{tabular}
		\caption{Comparison of different noise generators. Word deletion k\%: randomly mask k\% words.}
		\label{res-noise}
	\end{table}
		\begin{figure}[t]
		\centering
		\includegraphics[scale=0.20]{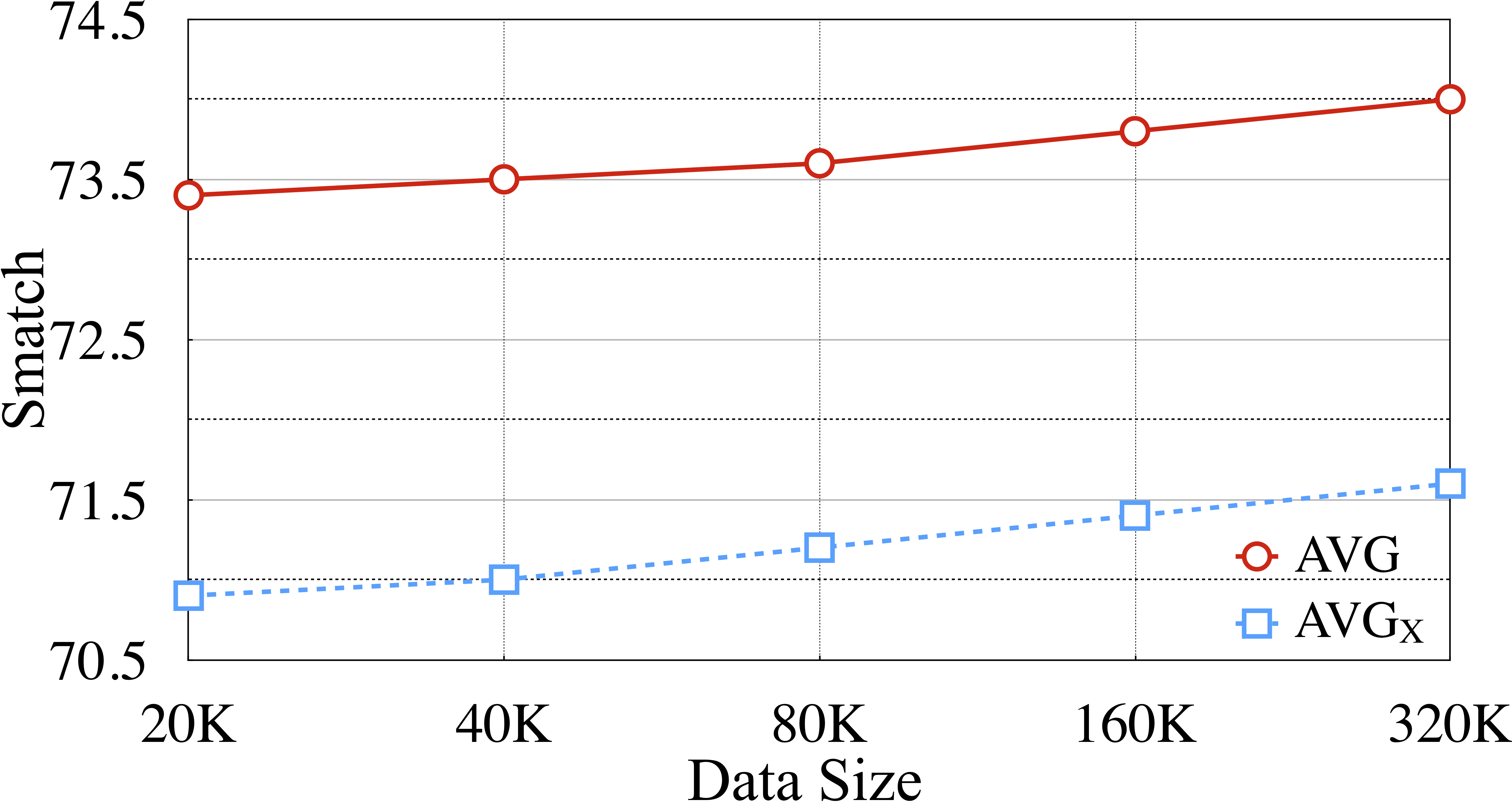}
		\caption{Comparison of different data sizes for KD.}
		\label{res-size}
	\end{figure}
	\paragraph{Effect of Data Sizes for Knowledge Distillation}
	Lastly, we study the relation between model performance and the size of monolingual data used for KD. Figure \ref{res-size} shows that the \textsc{Smatch} scores (AVG$_{\text{X}}$ and AVG) grow approximately logarithmically with the data size for KD.
    \section{Related Work}
    \paragraph{Cross-lingual AMR Parsing}
    AMR \cite{banarescu-etal-2013-abstract} is a semantic formalism initially designed for encoding the meanings of English sentences. Over the years, a number of preliminary studies have investigated the potential of AMR to work as an interlingua \cite{xue-etal-2014-interlingua,hajic-etal-2014-comparing,anchieta-pardo-2018-towards,zhu-etal-2019-towards}. These works attempt to refine and align English AMR-like semantic graphs labeled in different languages. \newcite{damonte-cohen-2018-cross} show that it is possible to use the original AMR annotations devised for English as representation for equivalent sentences in other languages and release a cross-lingual AMR evaluation benchmark \cite{damonte2020abstract} very recently. Cross-lingual AMR parsing suffers severely from the data scarcity issue; there is no gold annotated training data for languages other than English. \newcite{damonte-cohen-2018-cross} propose to build silver training data based on external bitext resources and English AMR parser. \newcite{blloshmi-etal-2020-xl} find that translating the source side of existing English AMR dataset into other target languages produces better silver training data. \newcite{sheth2021bootstrapping} focus on improving cross-lingual word-to-node alignment for training cross-lingual AMR parsers that rely on explicit alignment. Our work follows the alignment-free \textit{seq2seq} formulation \cite{barzdins-gosko-2016-riga,konstas-etal-2017-neural,van2017neural,peng-etal-2017-addressing,zhang-etal-2019-amr,ge2019modeling,bevilacqua-etal-2021-one} and we alternatively study this problem from the perspective of knowledge distillation, which provides a new way to enable multilingual AMR parsing.
	\paragraph{Knowledge Distillation for Sequence Generation}
	Knowledge distillation (KD) is a classic technique originally proposed for model compression \cite{bucilua2006model,ba2014deep,hinton2015distilling}. KD suggests training a (smaller) student model to mimic a (larger) teacher model, by minimizing the loss (typically cross-entropy) between the teacher/student predictions \cite{romero2015fitnets,yim2017gift,zagoruyko2017paying}. KD has been successfully applied to various natural language understanding tasks \cite{kuncoro-etal-2016-distilling,hu-etal-2018-attention,sanh2019distilbert}. For sequence generation tasks, \newcite{kim-rush-2016-sequence} first introduce sequence-level KD, which aims to mimic the teacher’s actions at the sequence-level. KD has been proved useful in a range of sequence generation tasks such as machine translation \cite{freitag2017ensemble,tan2019multilingual}, non-autoregressive text generation \cite{gu2017non,zhou2019understanding}, and text summarization \cite{liu2020noisy}. To the best of our knowledge, our paper is the first work to investigate the potential of knowledge distillation in the context of cross-lingual AMR parsing.
	\section{Conclusion}
	We presented a multilingual AMR parser that significantly advances the state-of-the-art parsing accuracies on multiple languages. Notably, the superior results are achieved with one single AMR parser. Our parser is trained with multiple pre-training and fine-tuning stages including a noisy knowledge distillation stage. We hope our work can facilitate the application of AMR in multilingual scenarios.
	\bibliography{anthology,custom}
	\bibliographystyle{acl_natbib}
\end{document}